\documentclass[journal]{IEEEtran}
\usepackage[utf8]{inputenc}
\usepackage{graphicx}
\usepackage{enumitem}

\usepackage[colorlinks=true, allcolors=blue]{hyperref}
\usepackage[colorinlistoftodos]{todonotes}

\usepackage{amsmath}
\definecolor{orange}{HTML}{FFC17D}
\definecolor{blue}{HTML}{7BABFF}
\definecolor{green}{HTML}{A1D68B}
\definecolor{lightgray}{HTML}{E8E8E8}

\usetikzlibrary{arrows.meta,shadows.blur}

\usepackage{adjustbox}
\usepackage{caption}    
\usepackage{tabularx}
\usepackage{makecell}                                                                                                                                                                               

\usepackage{authblk}

\usepackage{array}
\newcommand{\PreserveBackslash}[1]{\let\temp=\\#1\let\\=\temp}
\newcolumntype{C}[1]{>{\PreserveBackslash\centering}p{#1}}
\newcolumntype{R}[1]{>{\PreserveBackslash\raggedleft}p{#1}}
\newcolumntype{L}[1]{>{\PreserveBackslash\raggedright}p{#1}}

\begin{document}
\title{Explainable Artificial Intelligence (XAI) on Time Series Data: A Survey}

\date{} 


\author{Thomas~Rojat,
        Raphaël~Puget,
        David~Filliat,
        Javier~Del~Ser,
        Rodolphe~Gelin,
        and~Natalia~Díaz-Rodríguez
\thanks{T. Rojat is with U2IS, ENSTA Paris, Institut Polytechnique Paris and Inria Flowers and  Renault, France, France. Email: 
\href{mailto:thomas.rojat@renault.com}{thomas.rojat@renault.com}.}
\thanks{R. Puget is with Renault, France, France. email:
\href{mailto:raphael.puget@renault.com}{raphael.puget@renault.com}.}
\thanks{D. Filliat is with U2IS, ENSTA Paris, Institut Polytechnique Paris and Inria Flowers, France. Email: 
\href{mailto:david.filliat@ensta-paris.fr}{david.filliat@ensta-paris.fr}.}
\thanks{J. Del Ser is with TECNALIA, Basque Research and Technology Alliance (BRTA), 48160 Derio, Spain and the University of the Basque Country (UPV/EHU), 48013 Bilbao, Spain- Email: 
\href{mailto:javier.delser@tecnalia.com}{javier.delser@tecnalia.com}.}
\thanks{R. Gelin is with Renault France, France. Email:
\href{mailto:rodolphe.gelin@renault.com}{rodolphe.gelin@renault.com}.}
\thanks{N. Díaz-Rodríguez is with U2IS, ENSTA Paris, Institut Polytechnique Paris, Inria Flowers Team, 828 boulevard des Maréchaux, 91120 Palaiseau, France. Email: \href{mailto:natalia.diaz@ensta-paris.fr}{natalia.diaz@ensta-paris.fr}}}


\markboth{Under review}%
{Rojat \MakeLowercase{\textit{et al.}}: Explainable AI on Time Series Data}
\maketitle
\begingroup\renewcommand\thefootnote{\textsection}

\begin{abstract}

Most of state of the art methods applied on time series consist of deep learning methods that are too complex to be interpreted. 
This lack of interpretability is a major drawback, as several applications in the real world are critical tasks, such as the medical field or the autonomous driving field.
The explainability of models applied on time series has not gather much attention compared to the computer vision or the natural language processing fields. 
In this paper, we present an overview of existing explainable AI (XAI) methods applied on time series and illustrate the type of explanations they produce. We also provide a reflection on the impact of these explanation methods to provide confidence and trust in the AI systems. 

\end{abstract}



\IEEEpeerreviewmaketitle

\pagenumbering{gobble} 

\section{Introduction} \label{Introduction}


Time series are ubiquitous in nature, as they can represent any variable that varies over time. In industry, too, there are areas such as the medical field where temporal data is of particular importance. Thus, machine learning applied to temporal data is of particular importance as it has many potential applications in various fields. Several tasks can be performed by the machine learning methods applied to time series, the main ones being time series classification, time series forecasting, and time series clustering.
To perform these tasks, deep learning (DL) methods are since several years state of the art models for problems with time series as input data. Recurrent methods are adapted to work with time series thanks to their memory state and their ability to learn 
relations through time. Convolutional neural networks (CNNs) with temporal convolutional layers are able to build temporal relationships as well, and generate high level features from raw data. The introduction of those methods to work on time series enable to increase the accuracy of the models and avoid the heavy data pre-processing of former methods that could not directly take as input raw data. However, one of the major drawback of these methods is the lack of interpretability due to their high complexity. EXplainable Artificial Intelligence (XAI) is therefore a key concern for time series as most state of the art methods are not interpretable. This is a major drawback, as several applications of time series in the real world, such as the medical field or the autonomous driving field, are of critical importance and therefore require interpretability. 

Although a lot of work has been done on explainability in the computer vision and natural language processing (NLP) fields, there is still a lot of work to be done to explain methods applied on time series. This might be caused by the unintuitive nature of time series \cite{Siddiqui2019}, that we can not understand at first sight. Indeed, when a human being looks at a photo, or reads a text, he intuitively and instinctively understands the underlying information contained in the data. Although temporal data is ubiquitous in nature, through all forms of sound for example, humans are not used to representing this temporal data in the form of a signal that varies as a function of time. This may have an impact on the EXplainable Artificial Intelligence (XAI) applied to time series, especially for the evaluation of explanations, where qualitative assessments may have less potential than for the domains of computer vision or natural language explanation.
We need some expert knowledge or additional methods to leverage the underlying information present in the data. This problem might explain why, as for the computer vision field, a lot of existing methods focus on highlighting the parts of the input responsible for a prediction \cite{ribeiro2016i} \cite{Selvaraju_2019}.  


Motivated by the rationale above, the aim of this overview is to critically examine the current state of the art related to the explainability of models learned from time series data. Specifically, the contributions of this survey can be summarized as follows:
\begin{itemize}[leftmargin=*]
    \item An overview of the XAI methods applied on time series through their methodology, scope, and targets. 
     
     \item An overview of explainable methods that can be used to increase the confidence, stability and robustness of models.    
    
    \item An overview of the different approaches to qualitative and quantitative evaluate explanations provided by XAI methods applied to time series. 
    
    \item A discussion on the state of the art of explainability methods applied on time series, its limitations and potential lines of research.

\end{itemize}

The rest of the paper is organized as follows: First we present some important definitions of the XAI field, and the concepts of confidence, robustness and trustworthiness in Section \ref{Terminology}. Then, we present an overview of XAI methods applied on time series through their methodology in  \ref{Methodologies}, scope in Section \ref{Scope} and targets in Section \ref{Target}. Then, we tackle the evaluation of the explanations in the Section \ref{Evaluate}. 
Finally, Section \ref{Discussion} discusses the main outcomes of the paper. Figure \ref{fig:structure} illustrates schematically the overall structure followed in our survey.
\begin{figure}[h!]
    \centering
    \includegraphics[width=\linewidth]{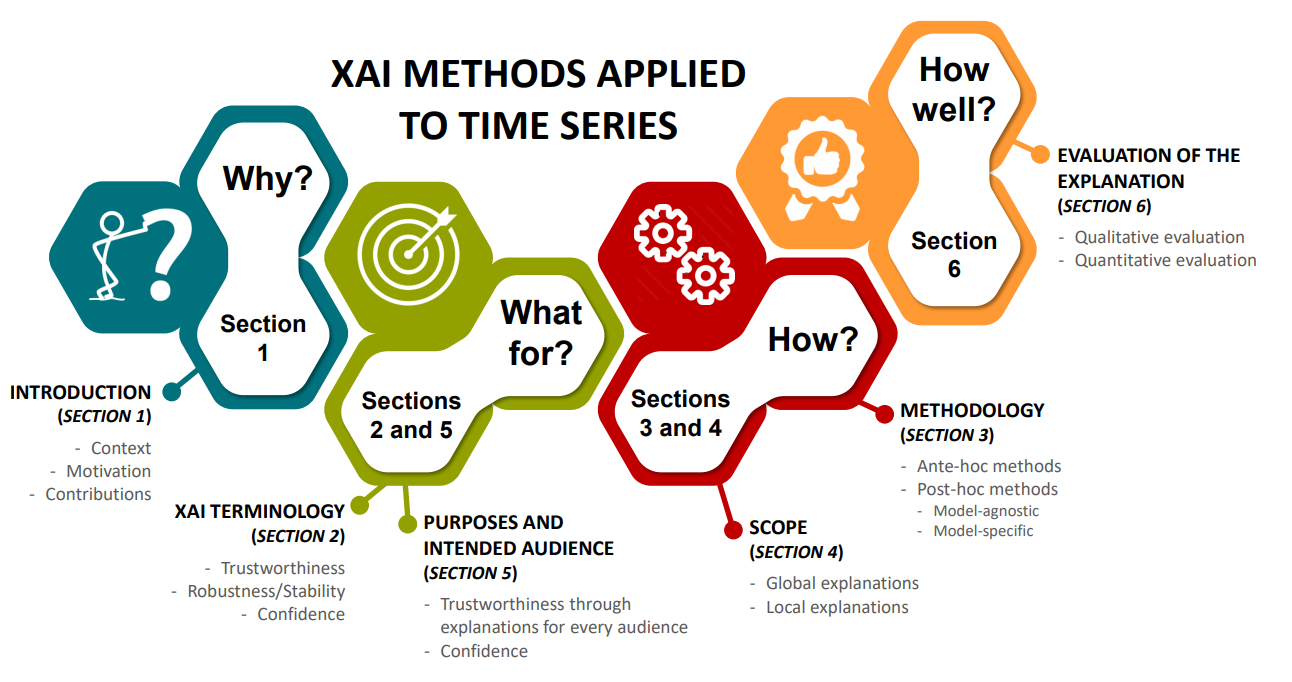}
    \caption{Questions answered throughout the survey and their connection to the different sections in which it is structured.}
    \label{fig:structure}
\end{figure}

\section{XAI Terminology and definitions} \label{Terminology}


The explainability methods always have one or more objectives to achieve through the explanations that generate. These goals will have their importance in the choice of the methodology, the scope of the explanations, and the targets. The following is a list of potential purposes of explainability methods that are necessary to know for understanding the rest of the paper, which we relate to each other in the knowledge graph depicted in Figure \ref{fig:kg}:
\begin{itemize}[leftmargin=*]
    \item \textbf{Explainability}: An "\emph{active characteristic of a model, denoting any action or procedure taken by a model with the intent of classifying or detailing its internal functions}" \cite{Tjoa_2020, arrieta2020explainable}. "\emph{Given an audience, an explainable Artificial Intelligence is one that produces details or reasons to make its functioning clear or easy to understand}" \cite{arrieta2020explainable}. 
    \item \textbf{Interpretability}: The passive characteristic of a model referring to the level at which a given model makes sense for a human observer \cite{Tjoa_2020}. An interpretable system is a system where a user cannot only see, but also study how inputs are mathematically mapped to outputs \cite{doran2017does}.  

    \item \textbf{Trustworthiness}: The "\emph{confidence of whether a model will act as intended when facing a given problem}" \cite{Tjoa_2020}. Trust can be achieved when the model can provide "\emph{detailed explanations}" of its decisions \cite{doran2017does}. A person may be more confident using a model if he understands it \cite{lipton2017mythos}.

    \item \textbf{Interactivity}: The interactivity with the user is "\emph{one of the goals targeted by an explainable machine learning model}" \cite{arrieta2020explainable}. This is specially important in fields where "users are of great importance". 
    
    \item \textbf{Stability}: A model is stable if it is not misled by small perturbations that might occur in the real world, such as noises due to the source of data itself (e.g. thermal noise in a sensor).
    
    \item \textbf{Robustness}: A model is considered robust if it is able to withstand disturbances that may have been intentionally created by humans.  
    
    \item \textbf{Reproducibility}: A model is reproducible if it repeatedly obtains similar results when run several times on the same dataset.
    
    \item \textbf{Confidence}: 
    The confidence is the probability of an event coming true.
    The goal is to quantify the trust in the decision \cite{kailkhura2019reliable}. It is defined in \cite{Wanner2020} "\emph{measure of risk as to how sure users are that they received the correct suggestions by the AI-based model}". A model with a high confident score on its predictions should be reproducible as it should get similar predictions when it is repeatedly run on the same dataset.  
\end{itemize} 
\begin{figure}[h!]
    \centering
    \includegraphics[width=0.85\linewidth]{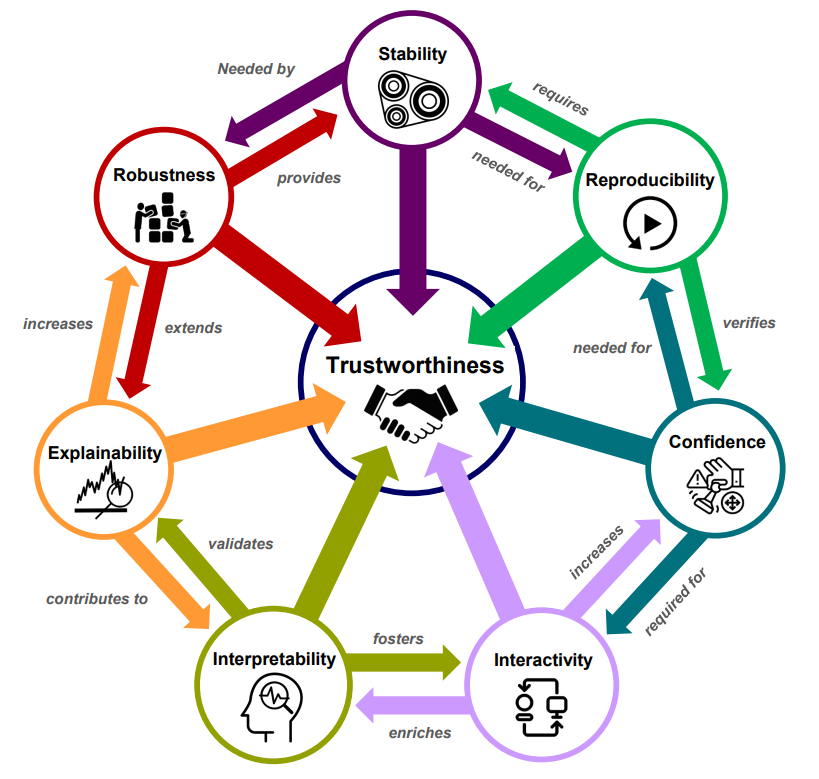}
    \caption{Knowledge graph relating all the purposes of explainability methods for time series.}
    \label{fig:kg}
\end{figure}

 The whole purpose of explainability is to explain models that are too abstract to be interpretable by themselves. 
 The need for explainability arises in certain practical scenarios when the task to perform is both too complex to be solved by a simple interpretable model, and too critical to be solved by a model that we cannot understand, and therefore cannot trust. 

Providing trustworthiness by explaining the inner behavior of these complex models can be one way to overcome these limitations. Most of the methods covered in this survey aim to provide trustworthiness in the model they explain. 
The interactivity with the final user is often neglected. Many methods provide insights on model behaviors without considering how a final user would receive the information provided by their methods.


The explainability brought by most of the methods presented in this survey can not assert the stability, robustness, and confidence in machine learning (ML) models applied on time series. This is why, to the best of our knowledge, there still are needs for developing metrics that will guarantee the right behavior of a model. 
Indeed there is little interest in providing an explanation if a small input noise can radically change the model behavior, making this explanation valid only for a very specific example. 

As shown in Figure \ref{fig:structure_term}, in the remainder of this section we elaborate on how XAI methods for time series analysis can contribute to the stability, robustness and confidence of systems with this particular kind of data at their core, delving into the contributions reported to date where such purposes have been targeted.
\begin{figure}[h!]
    \centering
    \includegraphics[width=0.8\linewidth]{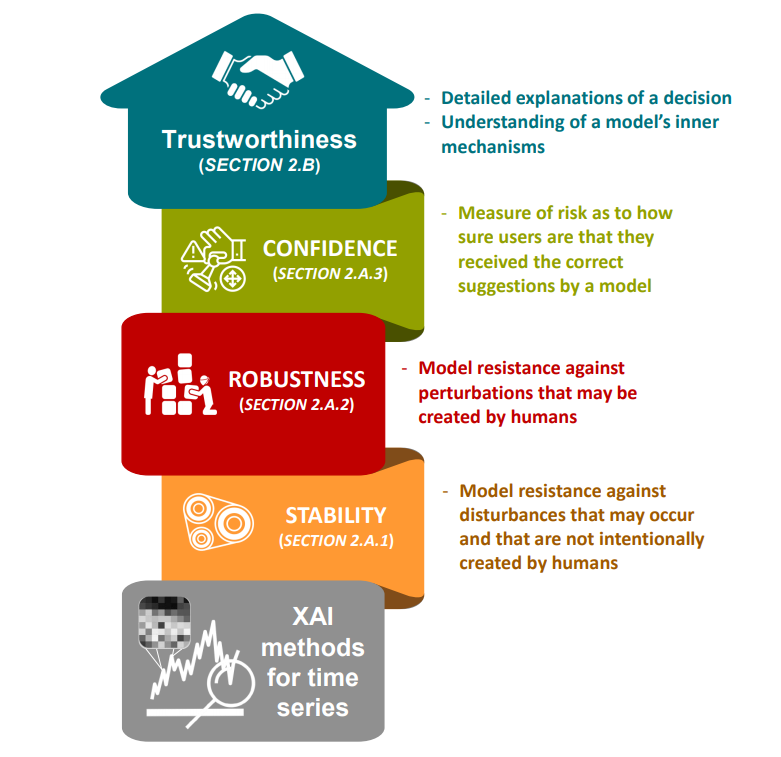}
    \caption{Rationale connecting the contents of Section \ref{Terminology}.}
    \label{fig:structure_term}
\end{figure}

\subsection{Stability, robustness and confidence of systems} \label{StabilityConfidenceRobustness}

It should be necessary for the certification of Artificial Intelligence (AI) systems to carefully assess the risks of the impacts of the Artificial Intelligence system on its environment \cite{hamonrobustness}. 
Let us take the example of an automated car. The risks involved are the road users that can be injured or killed if the automated system takes one wrong decision. The task to perform is critical and the risks associated are high. 

For a classification task, the metric always used to certify the quality of an AI system is the accuracy. It gives us the percentage of the examples that have been correctly classified, which is a satisfying insight to assess the quality of an AI system in most cases. However, some aspects are not covered by the accuracy. For instance, it cannot certify that the model output does not change when a small perturbation is applied on the input or the model, or when a noise is added in the input. Therefore, a high accuracy does not ensure the good behavior of the system, because such situations might occur in the real world. When the task to perform is critical, additional metrics are needed because we need to be sure that the AI system behaves well in every situation. 




The study of the stability, robustness, and confidence of AI system might be a way to tackle this limitations.

\subsubsection{Stability} \label{Stability}

As stated above, a model is stable if its output does not change when a small perturbation is applied on the input or on the model. These perturbations can occur in the real world. 

Let us take again the example of an automated car. Let us imagine that the car arrives in front of a stop sign. This stop sign is unusual because it has a white sticker sticked on its red part \cite{eykholt2018robust}. Eykholt et al. \cite{eykholt2018robust} show in their experiments that all the model deployed in the autonomous car missclassified these modified stop signs. However, it is crucial that the model identifies it correctly, because otherwise, it could lead to an accident. When facing this perturbation in the input sample, a stable model should be able to correctly classify the sample as a stop sign. If not, it should at least be able to warn the system that it is not certain of this prediction, as the situation encountered is unusual.



\subsubsection{Robustness} \label{Robustness}



A robust model is defined as a model that can withstand adversarial attacks.
Morgulis et al. \cite{morgulis2019fooling} conduct an experiment in which they perform adversarial attacks on traffic sign images by adding some imperceptible noise into the image such that human eyes cannot differentiate the real image and the modified one. They show that with the right amount of noise, the model deployed in the autonomous car is not able to identify correctly the objects in the modified images. The models that are fooled by these attacks cannot be labeled as robust. This might be a major security concern as some hackers can endanger the drivers life on the road simply by conducting these attacks on the input samples representing every type of traffic signs.

Another type of intentional perturbation of the inputs in order to modify the output of a given model are counterfactuals. Counterfactuals are defined in \cite{molnar2019} as \emph{"the smallest change to the feature values that changes the prediction to a predefined output"}. While perturbations generated by adversarial attacks are undetectable by humans, counterfactual perturbations are plausible and realistic because the modified samples are contained in the underlying distribution of data that can be encountered in the real world. 


\subsubsection{Confidence}


When an AI system encounters attacks and perturbations similar as those describe in Sections \ref{Stability} and \ref{Robustness}, we cannot have the guarantee that the model will remain stable and robust. There always might be a perturbation, or a noise that can mislead the system, and lead to a potential accident in the case of the automated car.  

Therefore, training the system to handle noises and perturbations might never be enough to guarantee the robustness and the stability of a system. However, what a model can potentially do is assess how unusual is a prediction vector compared to other points in the validation dataset \cite{xu2018interpreting}. It could identify examples that are far from the input data distribution, and thus identify samples that the model cannot classify with confidence. Thus, to every prediction could be associated a score relating how confident is the model in its decision. We could define a threshold, so that if the model confidence score is below this threshold, the model does not know and is not able to take a decision. 

Generally, uncertainty can be classified into two categories, aleatoric uncertainty and epistemic uncertainty. Aleatoric uncertainty is caused by the hazards that can occur when performing the same experiment several times and that can therefore change its result. Epistemic uncertainty is of different nature, it is due to limited data and knowledge. It occurs when a model encounters an example far from the distribution of input data, or when a model has difficulty extrapolating between the different examples in the learning base and thus generalizing. 

We will focus on epistemic uncertainty in this paper, and we will see in Section \ref{Confidence} how the explainability provided by XAI methods can be used to increase the robustness and the stability of a model by reducing epistemic uncertainty, and thus provide confidence in its outcomes. 


\subsection{User Trust }

Even if the explainability methods can bring confidence, explanations naturally bring trustworthiness thanks to the information they provide to better understand the model and its predictions.  
 While confidence and robustness may be approached with technical considerations, the trust of a user can also be built by other means than objective metrics. The trust of a user in an AI system can be both defined subjectively and objectively \cite{lipton2017mythos}. Interactions between the user and the AI system might be a key point to build trust between the automated system and the user \cite{haspiel2018explanations}. Without trust, users will not rely in automated systems, especially to conduct critical tasks. In automated driving for instance, AI systems suffer from the lack of trust of the users in the vehicle's autonomy \cite{petersen2017effects}. Such systems might suffer from a lack of feedback \cite{koo2015did} explaining for example why a specific action has been conducted by the automated system. These interactions are specifically important in semi-automated systems \cite{koo2015did}.

 The interactions are particularly important when the AI is in opposition to a user or a business stakeholder \cite{koo2016understanding}. The user might try to find why he disagrees with the AI prediction. He might therefore either find an example of failure of the system, or learn from its outcomes. This might not be achieved without proper feedback and interaction between the AI system and the user \cite{haspiel2018explanations}. Without it, it is rather unlikely that the driver handles the uncertainty and risk associated with giving driving control to the vehicle's autonomy \cite{robert2009individual}. 

One important concern is how to provide efficient interaction and feedback to the user \cite{koo2015did}. Providing the explanation before rather than after the action is likely to lead to greater trust in the autonomous system \cite{haspiel2018explanations}. Explaining why and how an action will be conducted might be appropriate to explain to users the action that is going to be conducted by the system \cite{koo2015did}. In semi-automated driving systems, providing drivers with an option to decide if the automated car will perform the action or not should lead to more trust beyond just providing an explanation \cite{haspiel2018explanations}. The feedback could also be from the human to the system, by giving the possibility to edit the system \cite{chander2018working} whenever the human identifies a failure. This would be an example of collaborative exploration that leads to improvement of both the user and the system. 

All the studies around the interactions between AI and humans should be human-centered \cite{koo2015did}. A stated in \cite{norman2016living}, "\emph{we must design our technologies for the way people actually behave, not the way we would like them to behave}". These involve the related notions of inclusion and accessibility \cite{diaz2020accessible,pisoni2021human}, related to the FAT and FATE AI (Fairness, Accessibility, Transparency, Ethics). The human responses to feedback and interaction with the system should be analyzed to progressively design interaction systems that lead to learn continuously \cite{lesort2020continual} and increase trust and acceptance of the system. 

Most of the explainable methods applied on time series produce explanations for the developers. Those methods focus on the technical aspects of their models without considering the human dimension of the explainability. More generally, for other types of users, ML is not the only field to consider when designing an explainable automated system. The psychology of the driver, for instance, is something to consider to be sure that the explanations provided will be useful for the user.

\section{
XAI techniques for time series} \label{Methodologies}

We will now present explainable methods that increase the trust in the ML model by explaining the prediction, or explaining what the model has learnt. All the methods we will present are applied on time series. First, we study post-hoc  methods explaining convolutional neural networks (CNNs). Post-hoc methods approximate the behavior of a model by extracting relationships between feature values and predictions \cite{Moradi_2021}. Post-hoc methods can be model-agnostic, usable on every type of models, or model-specific, only usable on one type of model. They are opposed to Ante-hoc methods that incorporates explainability into the structure of the model, that is thus already explainable at the end of the training phase.

The post-hoc methods that we will present are all specific to convolutional neural network (Table~\ref{table:TableMethods}) (Sections \ref{Backpropagation} and \ref{Perturbation}). They are all originally used in the computer vision field, and can be separated in two sections, back-propagation based methods and perturbation-based methods. Then, we will present some Ante-Hoc explainability methods specific to recurrent neural networks (RNNs), that might also be applied on the natural language processing (NLP) field, in Section \ref{Recurrent}. Finally, we will present some explainable data mining methods applied on time series in Section \ref{DataMining} and methods that provide explainability through representative examples in Section \ref{Representative}.   

\subsection{XAI for Convolutional Neural Networks}
\label{Convolutional}

We identify two types of methods to explain convolutional neural networks applied on time series, backpropagation-based methods and perturbation-based methods (Table~\ref{table:TableMethods}).

\subsubsection{Backpropagation-based methods} \label{Backpropagation}

Backpropagation methods provide explanations by doing a single forward and backward pass in the network. Most of the backpropagation-based explanation methods originally used to explain deep learning methods applied on images can also be used on DL methods applied on time series. 

Wang et al. \cite{wang2016time}, Fawaz et al. \cite{Ismail_Fawaz_2019}, and Oviedo et al. \cite{oviedo2018fast} use the class activation mapping (CAM) \cite{zhou2015learning}, a post-hoc method to provide explanations that  
highlights the regions in the input data that have the most influence on CNNs output classification prediction \cite{zhou2015learning}. CAM can highlight sub-sequences in the input time series that are maximally representative of a class. It relies on the presence of global average pooling layers at the end of the convolutional layers. The global pooling layer takes $N$ channels and returns its spatial average values. Channels with higher activations have higher signals.
Then, a weight is assigned per filter by using a dense linear layer with a softmax activation layer. Assuming there are $n$ classes, $n$ heatmaps are then created by computing the weighted sum of $N$ filters for every class. Finally, by up-sampling the class activation maps to the size of the input time series, we can identify the sub-sequences most relevant to any particular class. 
The class activation mapping method has the disadvantage that the model explained needs to have a specific architecture. A global average pooling layer needs to be added after the convolutional layers. 
Wang et al. \cite{wang2016time} conduct their experiments on the UCR time series repository datasets  \cite{chen2015ucr} with a little more than 80 datasets, which ares commonly used to evaluate models applied on time series. Fawaz et al. \cite{Ismail_Fawaz_2019} perform surgical skills evaluation using the JIGSAWS dataset \cite{gao2014jhu}, and Oviedo et al. \cite{oviedo2018fast} carry out time series classification from X-ray diffraction datasets. 


On the other hand, Strodthoff et al. \cite{Strodthoff_2019}, Siddiqui et al. \cite{Siddiqui2019} and Cho et al. \cite{cho2020interpretation} use the 'Gradient*Input' method which computes the partial derivative of the current layer with respect to the input and multiplies it by the input itself. Therefore, they compute the neurons and filters activation with respect to one specific instance. The input subsequences processed by the most activated filters have the highest contribution to the prediction (Fig.~\ref{fig:saliency}) \footnote[10]{For the sake of exemplifying the output of the reviewed techniques while acknowledging third party’s work, we include original figures after permission being granted by their corresponding authors.}.

\begin{figure}[htbp!]
   \centering
    \includegraphics[width=\linewidth]{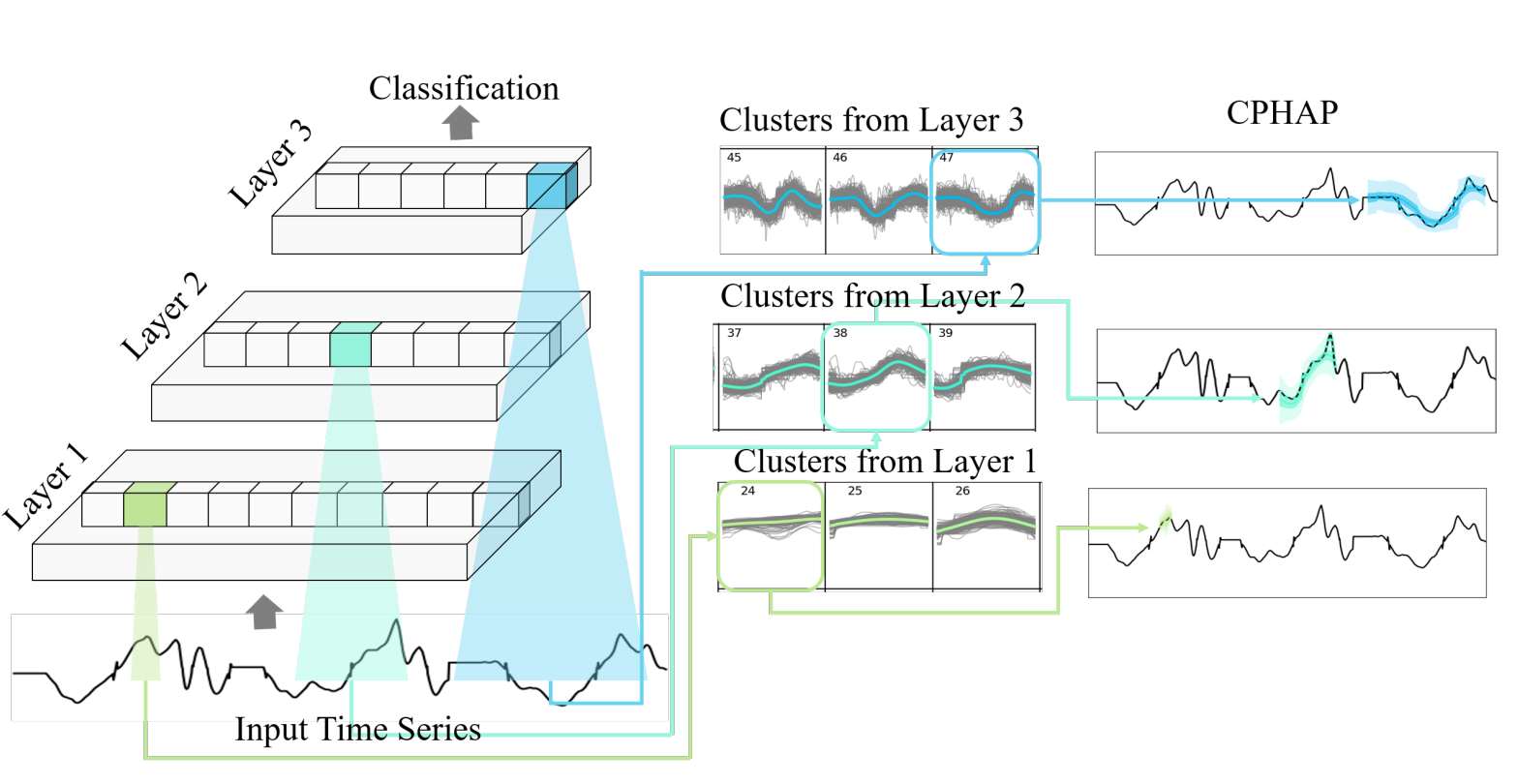}
    \caption{Usage of 'Gradient*Input' to identify the contribution of the input raw data when performing time series classification. It extracts the highly activated nodes in a channel and visualize the input sub-sequences that contribute to the highly activated nodes. Then, each extracted sub-sequence is assigned to a cluster of similar patterns. Figure reproduced with authorization from Cho et al. \cite{cho2020interpretation}.}
   \label{fig:saliency}.
\end{figure}

The 'Gradient*Input' approach can be used both for classification and regression tasks \cite{Siddiqui2019} as it only needs the neurons activations to produce its explanations. For the experiments, Strodthoff et al. \cite{Strodthoff_2019} detect myocardial infarction using an ECG dataset, the PTB diagnostic ECG dataset \cite{bousseljot1995nutzung,goldberger2000physiobank}. Siddiqui et al. \cite{Siddiqui2019} create a dummy dataset with threee features, the pressure, the temperature, and the torque to perform time series classification. Finally, Cho et al. \cite{cho2020interpretation} interpret deep temporal representations using two open source time series datasets: UWaveGestureLibraryAll \cite{liu2009uwave}, a set  of eight simple gestures generated from accelerometers, and Smartphone Dataset for human Activity Recognition \cite{asuncion2007uci}, a smartphone sensor dataset recording human perform eight different activities. 



\subsubsection{Perturbation-based methods} \label{Perturbation}

Perturbation-based methods directly compute the contribution of the input features by removing, masking, or altering them, running a forward pass on the new input, and measuring the difference with the original input \cite{ancona2018better}. The higher the difference, the higher the contribution of the input subsequence that has been altered. In theory, perturbation-based methods can be used as long as it is possible to compute distance values between the different outputs of the model. Thus, perturbation-based methods can be used for both classification and regression tasks \cite{tonekaboni2019explaining}. 

ConvTimeNet \cite{Kashiparekh2019} uses the occlusion sensitivity method~\cite{zeiler2013visualizing} which occludes parts of the time series and computes the difference in the probability $y$ for the predicted class (Fig.~\ref{fig:Kashiparekh2019}). They perform time series classification using 85 datasets taken from UCR \cite{chen2015ucr} TSC Archive Benchmark belonging from seven diverse categories: Image outline, Sensor Readings, Motion Capture, Spectrographs, ECG, Electric Devices and Simulated Data. 

Tonekaboni et al. \cite{tonekaboni2019explaining} define the importance of each observation as the change in the model output caused by replacing the observation with a generated one. They carry out mortality prediction with the Intensive Care Unit (ICU) Time Series dataset from MIMIC \cite{johnson2016mimic}.

\begin{figure}[htbp!]
    \centering
    \includegraphics[width=6.5cm]{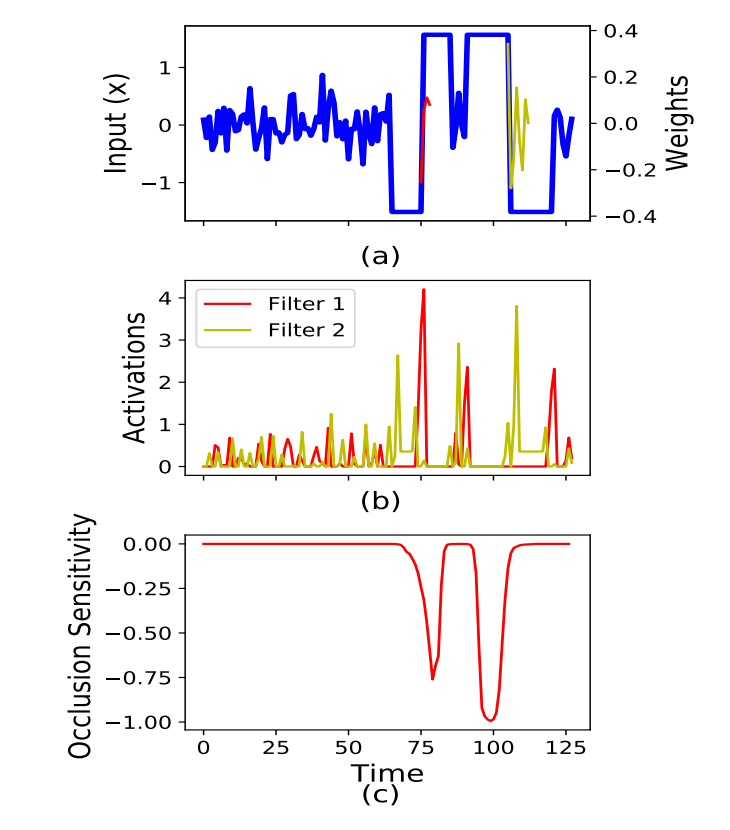}
    \caption{(a) Sample time series with top-2 relevant filters
from Two Patterns dataset, (b) their activation maps, and (c)
occlusion sensitivity plot. The goal is to use the occlusion sensitivity method \cite{zeiler2013visualizing} to compute the raw input contribution. Figure reproduced with authorization from Kashiparekh et al.~\cite{Kashiparekh2019}.}
    \label{fig:Kashiparekh2019}
    \vspace{-3mm}
\end{figure}


\subsection{XAI techniques for Recurrent neural networks} \label{Recurrent}

The convolutional neural networks are not the only deep learning methods that can perform time series classification. The recurrent neural networks, which are perfectly adapted to sequential data types, are also used to accomplish this kind of task. 

A first way to explain a recurrent model is to use attention mechanisms (Table~\ref{table:TableMethods}). Attention mechanisms assign values corresponding to the importance of the different parts of the time series according to the model  (Fig.~\ref{fig:schockaert2020attention}). It helps to overcome the fact that RNNs can't encode the information from too long input sequences. Attention mechanisms can be used for time series classification \cite{Karim2018} or time series forecasting \cite{schockaert2020attention}.

Choi et al. \cite{Choi2019} combine a CNN as a feature extractor and a Long Short Term Memory (LSTM) model to learn the temporal dependencies. Then, the hidden states and output states of the  Long Short Term Memory (LSTM) are used as input of a feedforward neural network layer which performs classification. The weights of this feedforward layer are the attention weights that indicate the importance of the different timesteps of the time series \cite{Karim2018,schockaert2020attention}. Choi et al. \cite{Choi2019} use the same feedforward layer to compute temporal attention, but they stack another neural network layer that takes as input the output of the temporal attention layer and the entire memory state of the LSTM to compute variable attention. Vinayavekhin et al. \cite{vinayavekhin2018focusing} compute more focused attention than the previous methods by using the whole input sequence to calculate an attention value for each timestep. Ge et al. \cite{ge2018interpretable} compute variable attention directly from the weights of the LSTM.  
To conduct their experiments, Schockaert et al. \cite{schockaert2020attention} generate an artificial dataset to perform time series forecasting for the temperature of the hot metal produced by a blast furnace. Vinayavekhin et al. \cite{vinayavekhin2018focusing} carry out time series classification using a public dataset for 3D human motion \cite{ionescu2013human3}. Finally,  Ge et al. \cite{ge2018interpretable} perform mortality prediction based on an ICU dataset.

On the other hand, attention mechanisms are also at the heart of transformers \cite{vaswani2017attention}, which can detect globally important variables for the prediction problem, persistent temporal patterns, and significant events that lead to significant changes in temporal dynamics \cite{lim2019temporal}. Lim et al. \cite{lim2019temporal} carry out time series forecasting on several real world datasets like the UCI Electricity Load Diagrams Dataset and the UCI PEM-SF Traffic Dataset.

To summarize, attention mechanisms are Ante-Hoc explainability methods (Table \ref{table:TableMethods}) because they are embedded in the structure of recurrent networks and the explicability they offer is available directly at the end of the learning phase. This is in opposition to the methods that explain convolutional neural networks (Section \ref{Convolutional}) which are specific post-hoc methods, as their explainability mechanisms are not incorporated in the structure of convolution networks, but are nevertheless only usable to explain convolution networks.
However, there is also the possibility to explain recurrent models by using a model-agnostic explanation method. Kim et al. \cite{Kim2019} use the SHapley Additive exPlanations (SHAP) algorithm \cite{lundberg2017unified}, a common model-agnostic feature attribution method, 
to explain the output of a recurrent model.

\subsection{Data mining based XAI models}
\label{DataMining}

As mentioned earlier, the deep learning explainability methods presented in the previous sections (Sections \ref{Convolutional} and \ref{Recurrent}) are not specific to the domain of time series data and can be applied to other fields. On the other hand, there are explainability methods only applicable to the time series. This is the case of the data mining methods that we will introduce in this section.
Several methods use data mining approaches to perform interpretable time series classification. Some of these methods are extensions of two data mining methods applied on time series: Symbolic Aggregate Approximation (SAX) \cite{Lin2007} (Table~\ref{table:TableMethods}) and Fuzzy Logic. 
\begin{figure}[h!]
    \centering
    \includegraphics[width=\linewidth]{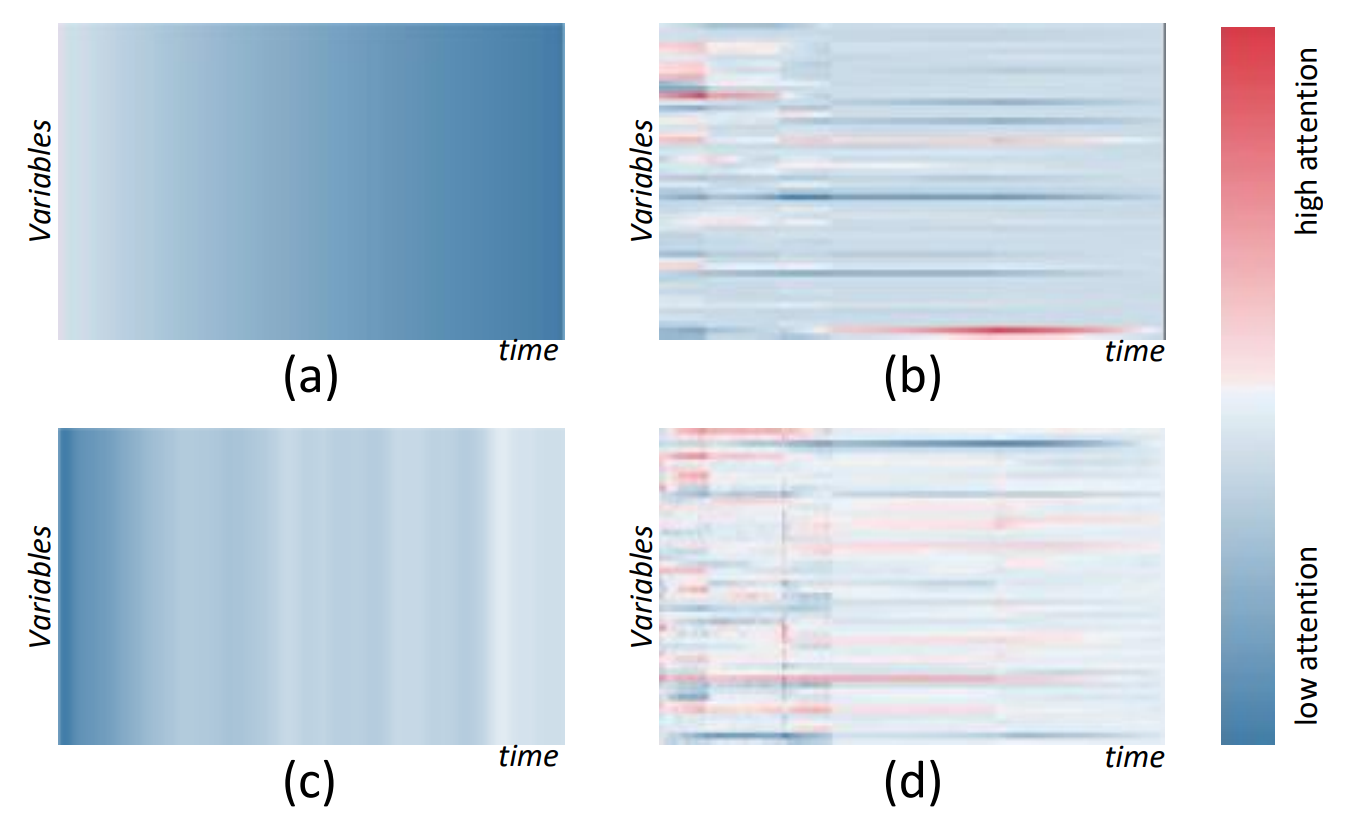}
    \caption{(a) Global temporal attention, (b) Global
spatio/temporal attention, (c) Local temporal attention, (d) Local
spatio/temporal attention. 'High attention' means a high contribution to the output while 'low attention' means a low contribution to the output. Attention  mechanisms  assign  values  corresponding  to  the  importance of  the  different  parts  of  the  time  series  according  to  the  mode. Figure reproduced with authorization from Schockaert et al.  \cite{schockaert2020attention}  
}
    \label{fig:schockaert2020attention}
\end{figure}

Symbolic Aggregate approXimation (SAX) \cite{Lin2003} transforms the input time series into strings. The algorithm consists of two steps. First, it transforms the time series into piece-wise aggregate approximation (PAA) representation \cite{Keogh2002}, and then converts this representation into strings. To transform the input data into piece-wise aggregate approximation (PAA) representations, the time series are split into equal-sized segments which are computed by averaging the values of these segments 
. Then, symbols are assigned to each segment. Assuming that the underlying input data distribution is Gaussian, each symbol is assigned to a segment by equiprobability with equal sized areas under the Gaussian curve. The input time series are then transformed into a sequence of symbols. This method is a well known way to detect recurrent patterns that occur in data.


Senin and Malinchik \cite{Senin2013}, and Le Nguyen et al. \cite{nguyen2018interpretableLearn} 
extend SAX to perform time series classification. 
They build interpretable high level features from raw data thanks to SAX, and select the best features according to each representations, providing both performance and interpretability. Compared to deep learning approaches, these approaches can be applied on variable-length time series, and are easier to interpret. To conduct their experiments, Senin and Malinchik \cite{Senin2013}, Le Nguyen et al. \cite{nguyen2020interpretable}, and Le Nguyen et al. \cite{nguyen2018interpretableLearn} perform time series classification using datasets from the UCR Time Series Classification Archive \cite{chen2015ucr}.

Fuzzy logic \cite{zadeh1988fuzzy}, fuzzy sets \cite{herrera2000fusion} and computing with words \cite{herrera2009computing} approaches are other methods used as drivers of explainability \cite{Mencar2019}. They aim at providing approximate reasoning and model outputs that are closer to natural language or use linguistic terms. In a way, their process is designed to be like human thinking \cite{Nayak2004}.
While in crisp rules, 
only 'True' or 'False' are accepted as outputs, in fuzzy logic outputs can be associated to any value between 0 and 1, giving a degree of possibility. It can be used to perform time series forecasting \cite{chen2004new,chen2000temperature} or detect hidden temporal patterns \cite{aydin2009prediction}. It can also be associated with neural networks to perform time series prediction \cite{Kasabov2002} and time series modeling \cite{Nayak2004}.


El-Sappagh et al. \cite{el2018ontology} and Wang et al. \cite{wang2020deep} combine the representational capacity of data-driven approaches and the intepretability of fuzzy-based approaches. To do that, El-Sappagh et al. \cite{el2018ontology} develop a fuzzy rule-based system (FRBS) to perform diabetes prediction from numerical time series data and static textual features. Wang et al. \cite{wang2020deep} propose a fuzzy cognitive map (FCM), a system with several components that can be connected to each other, and is interpretable because the interactions between components are weighted, to perform multivariate time series forecasting. They conduct their experiments on four real-world multivariate time series dataset. 

Paiva and Dourado \cite{paiva2004interpretability} also develop a neuro-fuzzy model to perform time series forecasting with the goal of solving complex tasks while keeping interpretability. That is why they use a linguistic model with fuzzy sets, instead of a Takagi-Sugano model with a first order fit function that is hard to interpret. They perform experiments on the Mackey-Glass time series \cite{mackey1977oscillation} and Box-Jenkis gas furnace datasets \cite{box2015time}.

\subsection{Explaining models through representative examples} 
\label{Representative}

Other types of methods, such as some methods that generate explanations by example, can be specific to the time series field. 
A type of explanations by example consists in giving the closest example in the training dataset that explains, as prototype, what the typical behaviour of a similar sample would look like. In this area we can find  embeddings-based models that look at the k-Nearest Neighbours (kNNs) in the embedding space to a given data point \cite{lee2012nearest,geler2020weighted,xu2009time}.

An example of methods that produce explanations by example specific to time series are \textit{Shapelets}, which are time series subsequences that are maximally representative of a class \cite{Ye2009} (Fig.~\ref{fig:Li2020}).

Shapelets were first introduced by Ye et al. \cite{Ye2011} in time series classification to overcome the limitations of state of the art time series classifiers. Shapelets are more interpretable, faster, and more accurate than k-Nearest Neighbours (kNN) \cite{Cover1967} which is a traditional approach to perform time series classification \cite{Lee2012}. They are computed by finding subsequences and associated thresholds that maximize the information gain when splitting the set of all subsequences into two classes following their distance to the candidate shapelet.



One limitation of shapelets is that there is a choice to make between efficient training and interpretability \cite{kidger2020generalised}. Wang et al. \cite{wang2019learning} and Kidger et al. \cite{kidger2020generalised} develop a regularization term that constrains the model to learn more interpretable shapelets. Another limitation is the computational time. Fang et al. \cite{Fang2018} use PAA \cite{Keogh2002} to discover candidate shapelets and thus reduce the computational time. Finally, Li et al. \cite{Li2020} develop a new way to find shapelets and perform time series classification which strongly reduces the computational time. 
\begin{figure}[h!]
    \centering
    \includegraphics[width=8cm]{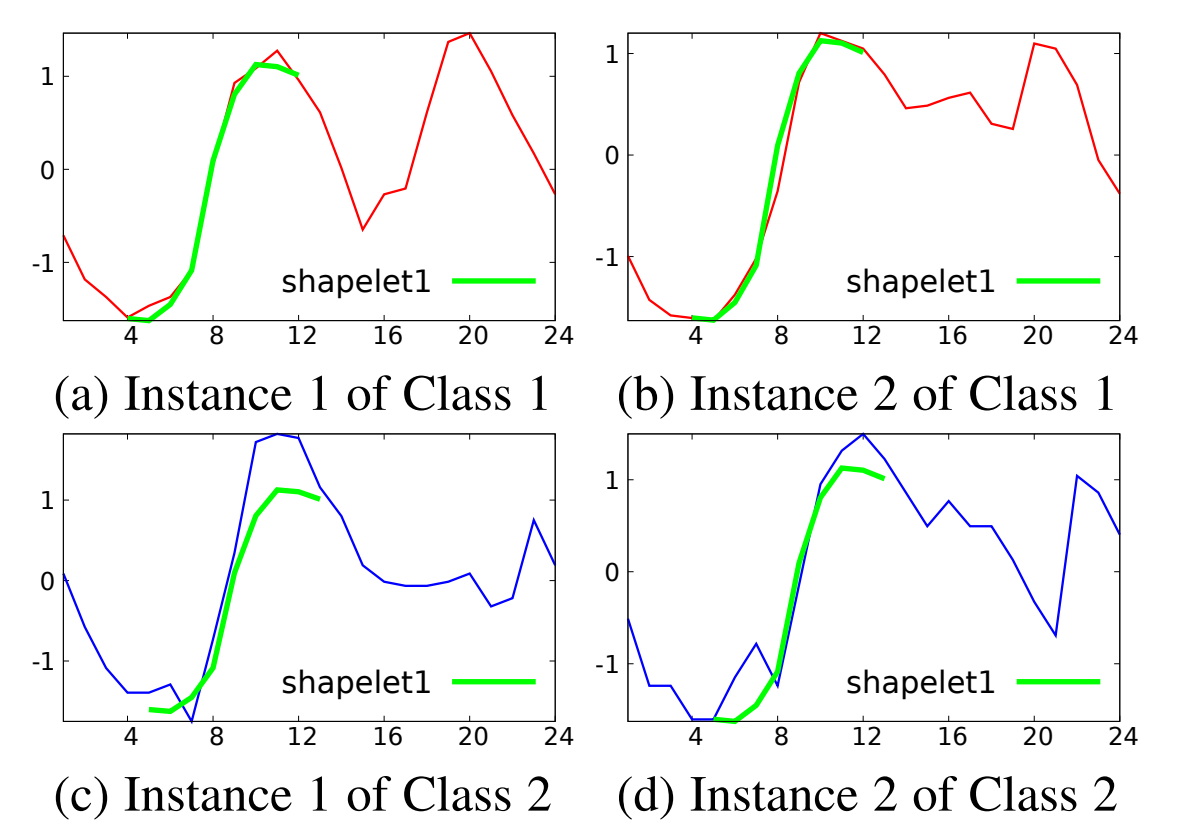}
    \caption{Example of \textit{shapelets}, explaining maximally representative subsequences of a class. Figure reproduced with authorization from Li et al. \cite{Li2020}.}
    \label{fig:Li2020}
\end{figure}

To conduct their experiments, Wang et al. \cite{wang2019learning}, Fang et al. \cite{Fang2018} and  Li et al. \cite{Li2020} carry out time series classification using datasets from the UCR repository Time Series Datasets \cite{chen2015ucr}, while Kidger et al. \cite{kidger2020generalised} perform time series calssification using datasets from the UEA Time Series Archive \cite{bagnall2018uea}.

\section{Explanations scale} \label{Scope}

The methods presented in the Section \ref{Methodologies} can provide local explanations or global explanations. The explanations are qualified as local when they are valid for a specific sample, and as global when they are valid for a set of samples or for the entire dataset. 


\subsection{Local explanations} \label{local}

Methods will tend to have local explanations when they make predictions sample by sample, and when the knowledge is not shared from one prediction to another. Thus, explainability methods specific to convolution networks naturally produce local explanations. 

Backpropagation-based methods (see Table~\ref{table:TableMethods}) rely on the activation of the neurons corresponding to a single prediction. Class Activation Mapping (CAM) \cite{zhou2015learning}, for instance, relies on the average of the channel activations of the last convolutional layer. The activation of the neurons change for every prediction and are therefore local parameters.

Perturbation-based methods such as ConvTimeNet \cite{Kashiparekh2019} (Table~\ref{table:TableMethods}) alter a sub-sequence from the input time series and look at the difference in prediction with the original sub-sequence. The computed relevances correspond to the sub-sequence that has been altered. 

Although, like convolutional neural networks, recurrent neural networks (Section \ref{Recurrent}) make their predictions sample by sample, they have a memory state that retains the knowledge built during previous representations. Unlike convolution neural networks, their latent representations are designed to handle one or several samples depending if the internal states are reset after each prediction. Therefore, the choice of this parameter will influence the scope of explanations of recurrent neural networks. The explanations are local if the internal states represent one instance, or global if the internal states represent several instances.

\subsection{Global explanations} \label{global}




Other methods like Shapelets (Section \ref{Representative}) or SAX (Section \ref{DataMining}) do not process the data sample by sample. For instance, the research of the candidate shapelet is not limited and can be carried on over the entire dataset. The scope of the explanations is defined by the size of the time series given as input.
\begin{figure}[h!]
    \centering
    \includegraphics[width=8cm]{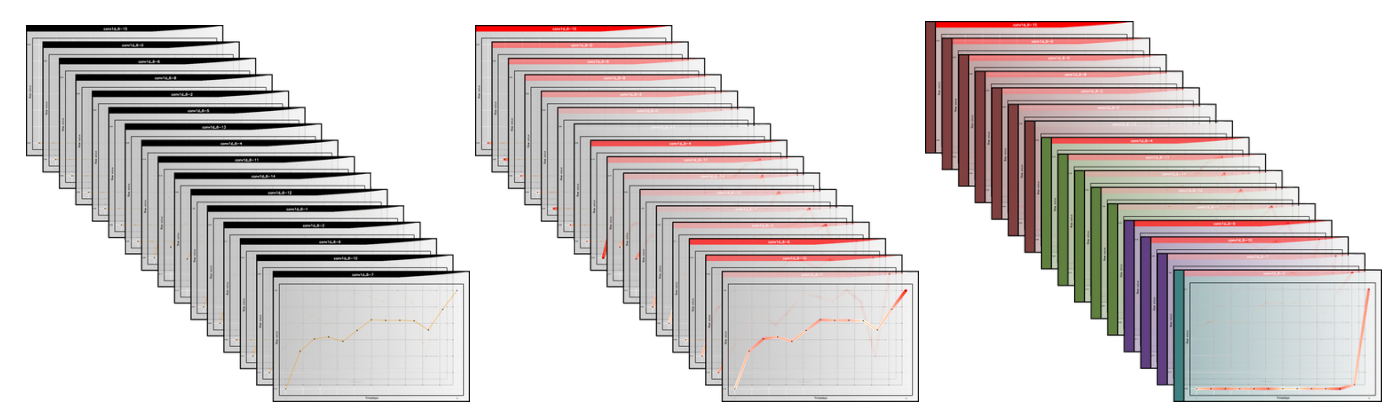}
    \caption{(a) 
    1-D Convolutional Filters, (b) 1-D Convolutional Filters with importance and saliency, and (c) 1-D Convolutional Filters with importance, saliency, and clusters. The goal is to compute input saliency and to cluster the filters. Figure reproduced with authorization from Siddiqui et al. \cite{Siddiqui2019}.}
    \label{fig:Siddiqui2019}.
\end{figure}

Finally, some papers extend the methods generating local explanations to produce global explanations. For instance, Oviedo et al. \cite{oviedo2018fast} generalize CAM to all training samples within a class. The average CAM allows to visualize the main discriminative features per class. 
Some methods provide global explanations through the usage of clustering. 
The backpropagation-based method Tsviz \cite{Siddiqui2019} highlights the important regions of the input data and computes the importance of filters for a given prediction. They also build global insights by clustering filters according to their activation pattern, as filters with similar activation patterns are essentially capturing the same concepts (Fig.~\ref{fig:Siddiqui2019}). Cho et al. \cite{cho2020interpretation} do not compute clusters with filters but with input sub-sequences. Each cluster is composed of a list of temporal sequences that activate the same nodes. They assign a general time series profile to these clusters with some uncertainty. This might be one of the most complete approach to explain a convolutional neural network with time series as input data. To the best of our knowledge, it is one of the only explainable methods applied on time series that explain the latent representations of convolutional neural networks.




\section{Purpose and intended audience of explanations} \label{Target}


The scope of explanations has an impact on the purpose of explainability methods. Global explanations can potentially be interesting to give trustworthiness and confidence, but local explanations may be more interesting for giving trustworthiness because their purpose is to explain the reasons for a prediction.

The scope of the explanations will also have an impact on the potential targets of explainability methods. Generally speaking, a target expert in machine learning, or a target with responsibility in the event of system failure, will rather aim at global explanations that will seek to explain the behaviour of the model as a whole. On the other hand, consumers of the model will rather look for local explanations that will explain the specific predictions that interest them.

First, we will present some applications of XAI methods providing trustworthiness for different types of targets. Then, we will show that some XAI methods are also able to increase the confidence in the model.  

\begin{table*}[htbp!] 
\centering
\resizebox{2\columnwidth}{!}{
\begin{tabular}{|C{3.0cm}|C{1.6cm}|C{3.0cm}|C{2.5cm}|C{1.8cm}|C{1.2cm}|C{2cm}|}  \hline
 \textbf{Model} &\textbf{Ante-hoc/} \textbf{Post-hoc} & \textbf{Methodology
 } & \textbf{Model Specific/} \textbf{Model Agnostic} & \textbf{Scope}  & \textbf{Target audience} & \textbf{Explanation evaluation}\\ \hline\hline
\textit{FCN}, Wang et al. \cite{wang2016time} &Post-hoc& Backpropagation-based &Specific& Local & Developer & No  \\ 
 \hline
 \textit{ESS-CNN}, Fawaz et al. \cite{Ismail_Fawaz_2019}&Post-hoc & Backpropagation-based &Specific& Local & User/DM & Yes/Qualitative   \\ 
 \hline
 \textit{XrayD-DNN}, Oviedo et al. \cite{oviedo2018fast}&Post-hoc & Backpropagation-based &Specific& Local/Global& Developer & Yes/Qualitative   \\ 
 \hline
 \textit{CY-EDL}, Wolanin et al. \cite{Wolanin2020}&Post-hoc & Backpropagation-based &Specific& Local/Global & User/DM & Yes/Qualitative   \\\hline
 
 \textit{Convtimenet}, Kashiparekh et al. \cite{Kashiparekh2019}&Post-hoc & Perturbation-based &Agnostic& Local & Developer & No    \\ 
 \hline
 \textit{FFC}, Tonekaboni et al. \cite{tonekaboni2019explaining}&Post-hoc & Perturbation-based &Agnostic& Local/Global & Developer & Yes/Quantitative   \\ 
 \hline
 \textit{MI-FCNN}, Strodthoff et al. \cite{Strodthoff_2019}&Post-hoc & Backpropagation-based &Specific& Local  & User/DM & No \\ 
 \hline
 \textit{Tsviz}, Siddiqui et al. \cite{Siddiqui2019}&Post-hoc & Backpropagation-based &Specific&Local/Global & Developer & Yes/Qualitative     \\  \hline
 \textit{CHAP}, Cho et al. \cite{cho2020interpretation}&Post-hoc & Backpropagation-based & Specific & Local/Global & Developer & Yes/Quantitative      \\ \hline
 \textit{CA-SFCN}, Hao et al. \cite{Hao2020}&Ante-hoc & Attention Mechanism &Specific& Local & Developer & No    \\ \hline
 \textit{ALSTM-FCN}, Karim et al. \cite{Karim2018}&Ante-hoc & Attention Mechanism &Specific& Local & Developer & No     \\ 
 \hline
 \textit{AM-LSTM}, Schockaert et al. \cite{schockaert2020attention}&Ante-hoc & Attention Mechanism &Specific& Local/Global & Developer & No         \\ 
 \hline
 \textit{TFT}, Lim et al. \cite{lim2019temporal}&Ante-hoc & Attention Mechanism &Specific& Local/Global & Developer & No        \\ 
 \hline
 \textit{Tsinsight}, Siddiqui et al. \cite{siddiqui2020tsinsight}&Ante-hoc & Attention Mechanism &Specific&Local/Global & Developer & No  \\ 
 \hline
 \textit{Sax-vsm}, Senin et al. \cite{Senin2013}&Ante-hoc & SAX &Specific& Global & Developer & No       \\ 
 \hline

 \textit{SAX-SFA-SEQL}, Le Nguyen et al. \cite{nguyen2018interpretableLearn}&Ante-hoc & SAX &Specific& Global & Developer & No\\ 
 \hline
 \textit{AI-PR-CNN}, Wang et al. \cite{wang2019learning}&Ante-hoc & Sapelets &Specific& Global & Developer & No   \\ 
 \hline
 \textit{BSPCOVER}, Li et al. \cite{Li2020}&Ante-hoc & Shapelets & Specific& Global & Developer & No       \\ 
 \hline
 \textit{GST}, Kidger et al. \cite{kidger2020generalised}&Ante-hoc & Shapelets &Specific& Global & Developer & No \\ 
 \hline

 \textit{eUA-CRNN}, Tan et al. \cite{Tan2020}&Ante-hoc & Attention Mechanism &Specific& Local  & Developer & No          \\ 
 \hline
 \textit{IDH-DSC-ERNN}, Choi et al. \cite{Choi2019}&Ante-hoc & Attention Mechanism &Specific& Local  & Developer & No        \\ 
 \hline
 \textit{ETNODE}, Gao et al. \cite{gao2020explainable}&Ante-hoc & Attention Mechanism &Specific& Local/Global  & Developer & No                \\ 
 \hline
 \textit{Tsxplain}, Munir et al. \cite{Munir_2019}&Post-hoc &  Backpropagation-based &Agnostic& Local  & Developer & Yes/Quantitative    \\ 
 \hline
  \textit{TCL}, Vinayavekhin et al. \cite{vinayavekhin2018focusing}&Ante-hoc & Attention Mechanism &Specific& Local  & Developer & Yes/Quantitative                 \\ 
 \hline
 \textit{ICU-LSTM}, Ge et al. \cite{ge2018interpretable}&Ante-hoc & Attention Mechanism &Specific& Global  & Developer & No                    \\ 
 \hline
 \textit{Series saliency}, Pan et al. \cite{pan2020series}&Post-hoc & Perturbation-based &Agnostic& Local  & Developer & Yes/Qualitative                     \\
 \hline
\textit{Mtex-cnn}, Assaf et al. \cite{Assaf2019}&Post-hoc & Backpropagation-based &Specific& Local & Developer & Yes/Quantitative \\
 \hline

 \textit{RATIO}, Augustin et al. \cite{augustin2020adversarial}&Post-hoc & Counterfactuals &Agnostic& Local & Developer &No \\
 \hline
 \textit{PDL}, Gee et al. \cite{gee2019explainin} & Ante-hoc & Prototypes &Specific& Global & Developer &Yes/Qualitative \\
 \hline
  \textit{FAHP}, El-Sappagh et al. \cite{el2018ontology} & Ante-hoc & Fuzy logic &Specific& Global & User &No \\
  \hline
 \textit{Deep-FCM}, Wang et al. \cite{wang2020deep} & Ante-hoc & Fuzzy logic &Specific& Global & Developer &Yes/Qualitative \\
\hline

\end{tabular}}

\captionof{table}{Summary of XAI methods applied to time series. Abbreviations:  }
 {SAX: Symbolic Aggregate Approximation; }{\label{tab:table-name}DM: Decision Maker; }

\label{table:TableMethods}
\end{table*}

\subsection{Providing trustworthiness through explanations for every audience}

In this section, we will present some applications of XAI methods applied on time series providing trustworthiness for three different type of targets: the developer, the decision maker, and the user. Some of the most interesting applications are in the medical field \cite{Ismail_Fawaz_2019}. Some other methods have interesting insights for model debugging \cite{oviedo2018fast,Strodthoff_2019}.

\subsubsection{Explanations for developers}

The developer is the one that makes the model. Most reviewed methods \cite{wang2016time,Kashiparekh2019,Kacem2018,tonekaboni2019explaining,Siddiqui2019,cho2020interpretation,siddiqui2020tsinsight} are not applied to any particular domain, but rather focus on an algorithm or a family of algorithms. 
As a consequence, the targets of these methods are mostly developers, as the insights provided by these methods are quite technical and hard to interpret for a non expert. 

However, Oviedo et al. \cite{oviedo2018fast} conduct their experiments on a specific domain but might be used by developers. The goal is to perform classification from a small x-ray diffraction time series dataset. They use explanations provided by CAM to identify the causes of correct and incorrect classifications. This outcome might only be interesting for developers .

Generally, developers seek for technical insights that can explain the whole model rather than explanations of a prediction. That is why developers are generally more interested in global explanations (Section \ref{global}) than local explanations (Section \ref{local}).




\subsubsection{Explanations for end-users}


A user is a person that consumes the output of the model. 
Let us take the example of the system that evaluates the surgical skills of young surgeons \cite{Ismail_Fawaz_2019}. The users here are the young surgeons that improve themselves by looking at discriminative behaviors specific to skill level. By identifying the gestures that made the model classify them as novices, they can identify their weaknesses and see how to improve themselves without the human intervention of an expert surgeon. 

Still in the medical field, clinicians can use the method that perform myocardial infarction detection \cite{Strodthoff_2019}. The interpretability is provided here to build confidence and trust in the model outcomes. Indeed, in a such critical fields, users cannot rely on the model if they do not know why the model takes the decisions. Thanks to the explanations provided, clinicians can check whether the model considers relevant certain patterns when making a decision.




Generally, users seek for the explanations of a prediction or a group of predictions that can affect them \cite{Ismail_Fawaz_2019,Strodthoff_2019}. Most of the time, they cannot take advantage of explanations of the whole model that may be hard to interpret and not necessarily fit to their situation. That is why users are generally more interested in local explanations (Section \ref{local}) than global explanations (Section \ref{global}).

\subsubsection{Explanations for decision makers}

The decision maker is characterized by a non-expertise, added to a liability in case of a problem with the AI system. For the method that evaluates the surgical skills of young surgeons \cite{Ismail_Fawaz_2019}, and for the method that performs myocardial infarction detection \cite{Strodthoff_2019}, the decision makers are the people in charge of the clinic. They are not experts in machine learning, but need guarantees that the system will work properly. They may not need an explanation of how the system works, but perhaps a measure of confidence (Section \ref{StabilityConfidenceRobustness}) in its decisions. In this way, they might know whether it is reasonable or too risky to use the system.

\subsection{Confidence} \label{Confidence}

The first purpose of explainable methods is to get insights on how the models work, how predictions are made. 
However, understanding how the model works is not enough. Particularly for critical tasks, we also need to be confident in the decisions taken by the model. We introduced in the Section \ref{StabilityConfidenceRobustness} the notion of confidence, and the potential link it can have with XAI. It is interesting to think that explainable methods might be useful to help models to be more robust and stable when facing samples out of the input distribution or adversarial attacks. We will present in this section some examples of XAI methods providing epistemic confidence in a model outcomes. Most of these methods are applied on time series data. 

Hartl et al. \cite{Hartl_2020} and Siddiqui et al. \cite{Siddiqui2019} 
use the raw features contribution to study adversarial attacks. Tsviz \cite{Siddiqui2019} introduce a perturbation on the most salient part of the input subsequence, leading to a huge drop in classification, which confirms the sensitivity of the model to noises. Hartl et al. \cite{Hartl_2020} develop a method named feature sensitivity that can quantify the potential of a feature to cause missclassification. Surprisingly, they found that the most salient features are the same than the features with highest potential to cause missclassification, and thus easily targetable by adversarial attacks. Therefore,  Hartl et al. \cite{Hartl_2020} propose a defense method in which they leave out the most manipulable features which does not lead to a huge drop of the accuracy. 

 Gee et al. \cite{gee2019explainin} introduce a diversity penalty to learn more diverse prototypes, which helps focusing on areas of the latent space where class separation is the most difficult. This helps the model to be more stable when classifying samples far from the input distribution. 
 
 Cho et al. \cite{cho2020interpretation} compare different attribution methods by perturbing the less salient input parts. The idea is that the neurons activations are more stable when we perturb the less salient input parts for the corresponding prediction. It is interesting to note that this is the opposite approach of Tsviz \cite{Siddiqui2019} which applies perturbations on the most salient parts of the input. However, both approaches \cite{cho2020interpretation,Siddiqui2019} are based on the same idea. Indeed, when perturbations are applied on the most salient parts of the input, we expect strong disturbances in the latent representations of the models \cite{Siddiqui2019}. Thus, when perturbations are applied on the less salient parts of the input, we expect small disturbances in the latent representations of the models \cite{cho2020interpretation}. 
 
The overlapping interpretable Sax words (Section \ref{DataMining} )
\cite{Senin2013} have individually very small contributions to the final prediction, which makes the methods less sensible to noises.   

Ates et al. \cite{ates2020counterfactual} combine two adversarial training procedures, Adversarial Training (AT) \cite{madry2019deep} and  Adversarial confidence enhanced training (ACET) \cite{hein2019relu}. Adversarial Training (AT) is better at making the model robust against adversarial attacks, while Adversarial confidence enhanced training (ACET) is better at tackling out of distribution samples. Ates et al. \cite{ates2020counterfactual} develop RATIO, a method that generates counterfactual \cite{Stepin2021} visual explanations. These counterfactual explanations visually show the changes in the input sample to reach the targeted AT and ACET confidence score.


Although this is only the beginning, these are to the best of our knowledge the first attempts to provide epistemic confidence in the models using the insights provided by XAI methods. 

\section{Evaluating explanations} \label{Evaluate}

After presenting the scope and the targets of explainability methods, we now tackle the evaluation of these XAI methods. There is not a metric globally recognized that can assess the quality of explanations (Table \ref{TableEvaluation}). This might be due to the different nature of the explanations generated and the different input data types. However, some quantitative evaluation approaches exist to objectively assess the quality of the explanations generated in several fields, including the time series domain. Qualitative evaluations can also be made by experts to assess the relevancy of the explanations generated. 

\subsection{Qualitative evaluations}

While some methods \cite{nguyen2020interpretable,Ismail_Fawaz_2019,oviedo2018fast,Karim2018,schockaert2020attention,Wolanin2020} do not perform any evaluation of the generated explanations, these methods could be evaluated using domain expert assessments. Global explanations provided by, for instance, average CAM in Oviedo et al. \cite{oviedo2018fast} could be assessed by experts by just analysing the global attribution maps computed by averaging every class activation map representing each class, and not all the local attribution maps one by one. Methods providing explanations for specific users \cite{Ismail_Fawaz_2019,Wolanin2020}  can also easily benefit from experts feedback. For example, the surgeons experimenting the model that performs evaluation of surgical skills \cite{Ismail_Fawaz_2019} can give some feedback, whether they found the explanations provided relevant or not. 

The backpropagation-based approach Tsviz \cite{Siddiqui2019} assesses the quality of generated explanations by analyzing its explicitness. The explicitness is provided by the clustering of hidden representations and showing the influence of these hidden representations on the output.  



However, unlike in computer vision, qualitative evaluations might have a limited potential in the time series field \cite{arnout2019towards}. Indeed, the unintuitive nature of time series (see Section \ref{Introduction}) make it difficult even for domain experts to qualitatively assess the quality of the explanations generated. Therefore, according to Arnout et al. \cite{arnout2019towards}, we should prioritize quantitative evaluations for the time series field.  

\subsection{Quantitative evaluations} \label{QuantEvaluations}

Arnout et al. \cite{arnout2019towards} presents a way to evaluate explanations providing the most contributing regions of the input time series to the model. They propose to conduct perturbation on the data, with the idea that if relevant features get changed, the performance of an accurate model should decrease massively. Tonekaboni et al. \cite{tonekaboni2019explaining} takes another approach by comparing its explanations with several state of the art feature attribution methods, sensitivity analysis \cite{bach2015pixel} \cite{Yang2018}, feature occlusion \cite{suresh2017clinical}, augmented feature occlusion \cite{suresh2017clinical}, and Local Interpretable Model-agnostic Explanations (LIME) \cite{ribeiro2016i}. Cho et al. \cite{cho2020interpretation} also assess their explanations by providing some comparisons with another attribution method, Layer-wise Relevance Propagation (LRP) \cite{binder2016layer}, thanks to some perturbation analysis:
they gradually alter subsequences that have been identified by the attribution methods as important for a given prediction. They compare both methods through the absolute difference of activations of neurons depending on the perturbation magnitude of their identified explanations (Fig.~\ref{fig:cho2020interpretation}). The perturbation-based method proposed by Tonekaboni et al. \cite{tonekaboni2019explaining} performs sanity checks by doing data and model randomization: they evaluate the faithfulness in Tsviz \cite{Siddiqui2019} by removing the filters that have the highest importance and check if the prediction changes.  
\begin{figure}[h!]
    \centering
    \includegraphics[width=\columnwidth]{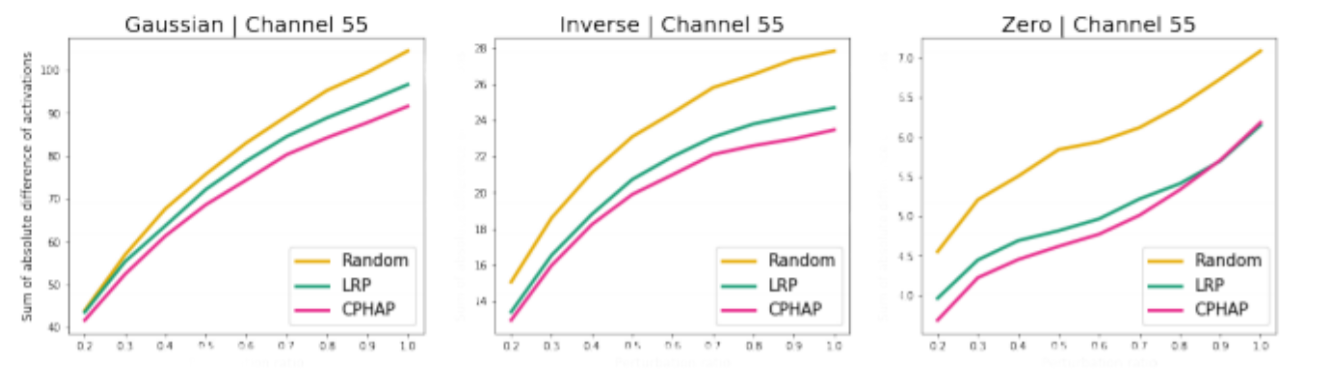}
    \caption{  This graph illustrates the results of perturbations on the input while preserving the regions
selected by each method. We apply three kinds of perturbations: Gaussian perturbation, Inverse
perturbation and zero perturbation. The x-axis means the ratio of perturbed regions except preserved
regions. The y-axis means the sum of changes of activations in the channel 55. Figure reproduced with authorization from Cho et al. \cite{cho2020interpretation}. 
}
    \label{fig:cho2020interpretation}.
\end{figure}

Finally, Arnout et al. \cite{arnout2019towards} propose two new quantitative evaluation methods to overcome the limitations of the perturbation approach, which has a lack of evaluation of trends or patterns in the time series \cite{arnout2019towards}. They propose two new sequence evaluation approaches, Swap Time Points and Mean Time Points, that take the inter-dependency of points into account. Swap Time Points invert the order of the points in the most salient sub-sequences, and compare the results with another sequence where the values of the corresponding sub-sequences have been put to 0. Mean Time Points has a similar approach, but instead of swapping the time points, it assigns the mean of all values of the salient sub-sequence to all points in the corresponding sub-sequence. These two approaches are complementary with the perturbation approach in the sense that they offer a new way to evaluate time series salient explanations, as they can evaluate trends or patterns of the time series.

\begin{table*}[htbp!]
\centering
\begin{adjustbox}{angle=0}
\begin{tabular}{|p{5.0cm}|p{5.0cm}|p{5.0cm}|}  \hline

\textbf{Model} &\textbf{Quantitative/Qualitative Evaluation} & \textbf{Evaluation Approach
 } \\ \hline\hline
\textit{REXAITS}, Arnout et al. \cite{arnout2019towards}& Quantitative & Perturbation approach/Sequence evaluation approach \\ 
 \hline
\textit{DADNN}, Oviedo et al. \cite{oviedo2018fast}& Qualitative & Expert assessment \\ 
 \hline
 
\textit{ESSCNN}, Fawaz et al. \cite{Ismail_Fawaz_2019}&Qualitative & Expert assessment  \\ 
 \hline
  
\textit{EUEDL}, Wolanin et al. \cite{Wolanin2020}&Qualitative & Expert assessment  \\ 
 \hline
   
\textit{ETSC}, Tonekaboni et al. \cite{tonekaboni2019explaining}& Quantitative & Perturbation approach/Sanity check \\ 
 \hline

\textit{Tsviz}, Siddiqui et al. \cite{Siddiqui2019}&Qualitative/Quantitative&Expert Assessment/Sanity check     \\  \hline
\textit{SVIAN}, Cho et al. \cite{cho2020interpretation}&Quantitative & Perturbation approach     \\ \hline

 \textit{Tsxplain}, Munir et al. \cite{Munir_2019}& Quantitative & Sanity check    \\ 
 \hline
  \textit{Focusing on what is relevant}, Vinayavekhin et al. \cite{vinayavekhin2018focusing}& Qualitative & Expert assessment                 \\ 
 \hline

 \textit{Series saliency}, Pan et al. \cite{pan2020series}& Qualitative & Sanity check                     \\
 \hline
\textit{Mtex-cnn}, Assaf et al. \cite{Assaf2019}& Quantitative & Perturbation approach \\
 \hline

 \textit{WKCED}, Gee et al. \cite{gee2019explainin}& Qualitative & Expert assessment \\
 \hline

\end{tabular}
\end{adjustbox}

\captionof{table}{Summary of evaluation approaches for XAI methods applied to time series.  }


\label{TableEvaluation}
\end{table*}


To summarize, we can reflect on the fact that depending on what we want to accomplish with the explanations, qualitative or quantitative evaluations may be more or less appropriate. Qualitative assessments may be better suited to explanations that target users or decision makers \cite{Ismail_Fawaz_2019}, while quantitative assessments may be better suited to explanations that attempt to discern new predictive knowledge in the data \cite{cho2020interpretation}. Then, at this level, a question comes to mind. Are these explanations that teach experts how the models work sufficient to develop the confidence to put the models into practice to accomplish critical tasks? We will attempt to answer this question in the next section (Section \ref{Discussion}).

\section{Discussion} \label{Discussion}

Some of the methods presented in this survey are originally applied to other areas than time series. The back-propagation based (Section \ref{Backpropagation}) and perturbation-based approaches (Section \ref{Perturbation}) were first designed for the computer vision field and then applied to the time series field. 
To the best of our knowledge, there is a lack of explainable methods applied on CNNs specifically designed for time series tasks. There must be specificities in time series data that can be exploited to design explainable approaches specific to CNNs that are uniquely adapted to the time series field. 

Most methods presented in this survey indicate which specific regions of the input data get attention from the model while classification is performed. They do not provide any confidence in the model, neither mitigates its vulnerabilities. However, it can be one way to provide some explanations and increase the trust in the system. Indeed, a user or a developer can rely more on the system if he knows that the model gives its attention to the relevant parts of the input for a specific prediction. However, the trust brought by these methods can be questioned by the unintuitive aspect of time series. For instance, saliency maps on images are directly interpretable as we usually understand the content. It is different for time series because expert knowledge might be needed to understand its outcomes. Some data mining methods might be useful to automatically extract the underlying content of time series. 

As highlighted in this survey, it is possible to use the insights provided by some explainable methods to increase the epistemic confidence in the model. 
Obviously, the purpose of XAI is to get some information and understanding of the model, and therefore to provide trust. However, the XAI field has more potential than just facilitating trustworthiness. Explainability has the potential to lead to new metrics and training practices ensuring the confidence and robustness of the most complex and abstract models, thanks to the insights explainable methods can provide. 

We are also far from getting an end-to-end XAI system, as methods only focus on the technical parts. XAI techniques do not consider interactions with the user or developer that are necessary for the AI system to be trusted and used. There is a lack of objective tools to demonstrate the robustness of AI systems. This is why interactive systems providing explanations and feedback might be a leading way to empirically and subjectively show the user and decision maker that the AI system can be trusted.

\section*{Acknowledgements}

We acknowledge the authors of \cite{wang2016time}, \cite{schockaert2020attention}, \cite{Kashiparekh2019}, \cite{Siddiqui2019} and \cite{Li2020} for letting us use their original figures for illustrative purposes. T. Rojat would also like to thank ANRT and Renault for the funding support. J. Del Ser would like to thank the Basque Government for its funding support through the EMAITEK and ELKARTEK programs (3KIA project, KK-2020/00049), as well as the consolidated research group MATHMODE (ref. T1294-19). 



\bibliographystyle{abnt-num}

\end{document}